\def\year{2020}\relax
\documentclass[letterpaper]{article} 
\usepackage{aaai20}  
\usepackage{times}  
\usepackage{helvet} 
\usepackage{courier}  
\usepackage[hyphens]{url}  
\usepackage{graphicx} 
\usepackage{graphicx}  
\usepackage{latexsym}
\usepackage{url}
\usepackage{booktabs}
\usepackage{multirow}
\usepackage{amsmath}
\usepackage{amsfonts}

\newcommand{\citet}[1]{\citeauthor{#1}~\shortcite{#1}}
\newcommand{\citep}{\cite}

\title{Bridging Text and Video: A Universal Multimodal Transformer \\ for Video-Audio Scene-Aware Dialog}
\author{ 
\textbf{Zekang Li$^{14}$, Zongjia Li$^{23}$, Jinchao Zhang$^2$, Yang Feng$^1$\thanks{Joint work with Pattern Recognition Center, WeChat AI, Tencent Inc, China. Yang Feng is the corresponding author. This work was done when Zongjia Li was interning at Pattern Recognition Center, WeChat AI, Tencent.}, Cheng Niu$^2$, Jie Zhou$^2$}\\
\textsuperscript{\rm 1}Key Laboratory of Intelligent Information Processing\\ 
Institute of Computing Technology, Chinese Academy of Sciences \\
\textsuperscript{\rm 2}Pattern Recognition Center, WeChat AI, Tencent Inc, China\\
\textsuperscript{\rm 3}School of EECS, Peking University \\
\textsuperscript{\rm 4}University of Chinese Academy of Sciences \\
{\tt \small \{lizekang19g, fengyang\}@ict.ac.cn, zongjiali@pku.edu.cn}\\
{\tt \small \{dayerzhang, niucheng, withtomzhou\}@tencent.com}
}
\begin{document}

\maketitle

\begin{abstract}

\end{abstract}
Audio-Visual Scene-Aware Dialog (AVSD) is a task to generate responses when chatting about a given video, which is organized as a track of the 8$^{th}$ Dialog System Technology Challenge (DSTC8). 
To solve the task, we propose a universal multimodal transformer and introduce the multi-task learning method to learn joint representations among different modalities as well as generate informative and fluent responses. Our method extends the natural language generation pre-trained model to multimodal dialogue generation task. Our system achieves the best performance in both objective and subjective evaluations in the challenge.

\section{Introduction}
\begin{figure}[t]
    \centering
    \small
    \includegraphics[width=0.5 \textwidth]{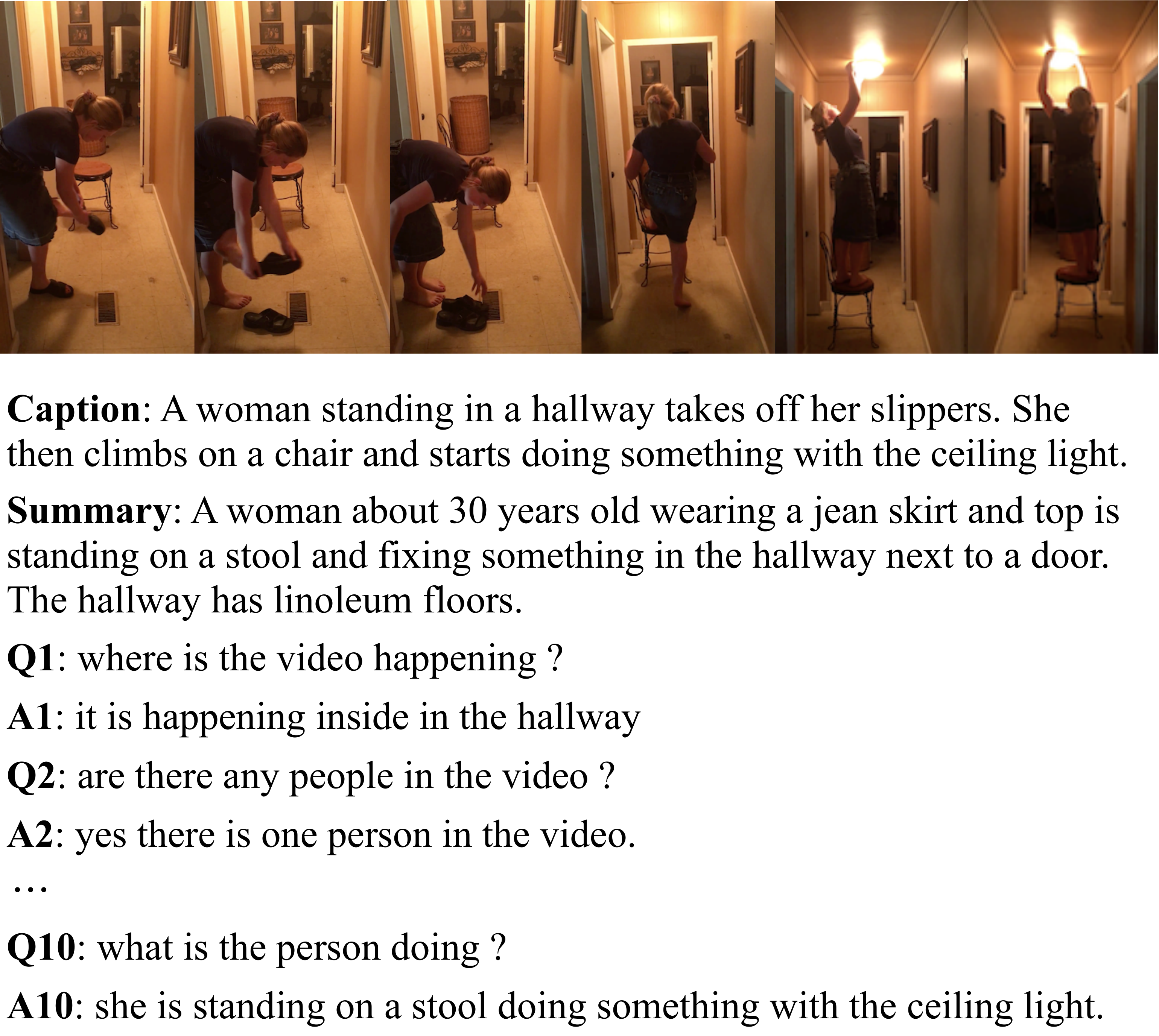}
    \caption{A dialogue sampled from the DSTC8-AVSD dataset. For each dialogue, there are video, audio, video caption, dialogue summary and 10 turns of conversations about the video.}
    \label{fig:data_example}
\end{figure}

Recently, scene-aware dialogue generation has attracted increasing attention in both industry and academia due to its broad application. It aims to generate informative and fluent dialogue responses grounding on the given scenes. 
\citet{zhou2018dataset} propose a dataset for text-based movie grounded conversations. 
\citet{urbanek2019light} builds a large-scale text adventure game platform, in which agents can act and speak grounded on the scenes described in the text. Inspired by human inherent multimodal understanding ability, \citet{alamri2018audio} integrates multimodality to scene-aware dialogue and proposes the Audio-visual Scene-Aware Dialog task. 

The goal of the Audio-Visual Scene-Aware Dialog task is to generate correct and fluent responses by understanding all modalities (e.g., text, video and audio), which is a more challenging task than image-based or text-grounded dialog tasks. Figure \ref{fig:data_example} shows an example dialogue in DSTC8-AVSD dataset \cite{alamri2018audio}. There are three challenges in this task: acquiring accurate representation of the video, effective interaction among different modalities and better understanding dialogues and generating responses. 
\citet{hori2019end} introduces an LSTM-based encoder and decoder with the multimodal attention. \citet{dat2019film} propose a hierarchical recurrent encoder-decoder framework for encoding and generating responses based on a FiLM-based audio-visual feature extractor. \citet{ramakanth2019scene} adopts a dual attention mechanism to encode and align multiple modalities. The winning team of the DSTC7-AVSD task \cite{sanabria2019cmu} focuses on using a hierarchical attention to combine textual and visual modalities and employ the How2 dataset for pre-training. Moreover, MTN \cite{le-etal-2019-multimodal} proposes multimodal transformer networks to encode video and incorporate information from different modalities.
 
\begin{figure*}[t]
    \centering
    \small
    \includegraphics[width=0.98 \textwidth]{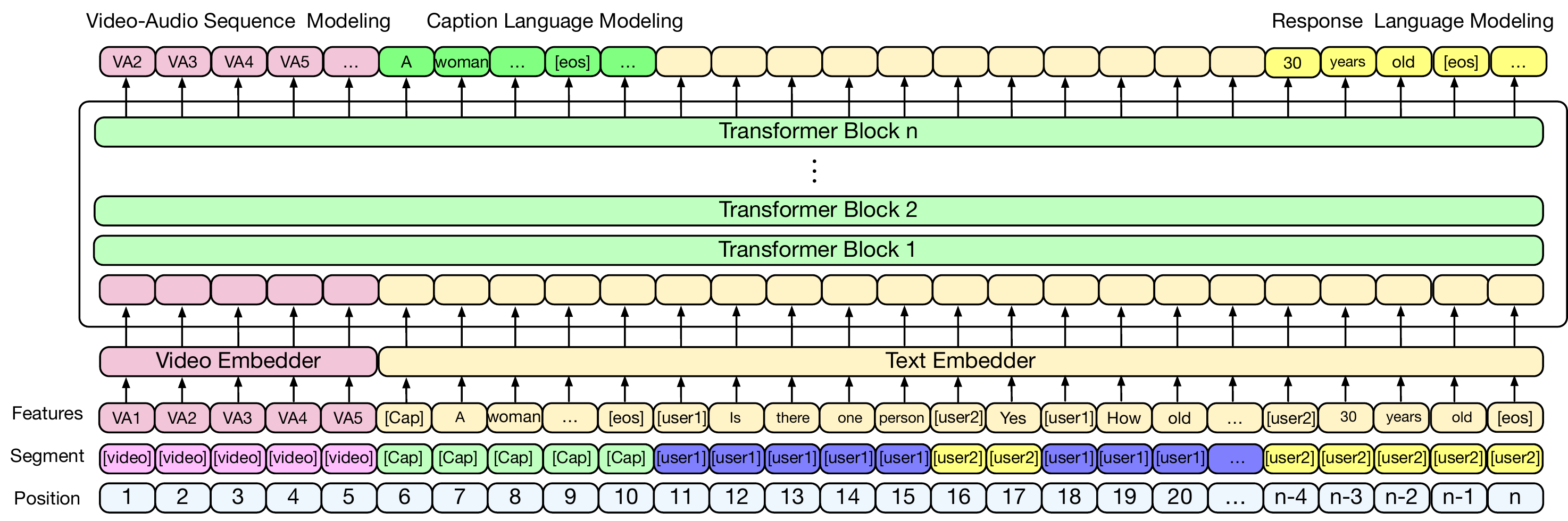}
    \caption{Our universal multimodal transformer architecture. We concatenate video-audio, caption, dialogue history, and response features to a long sequence. For different types of input, we adopt different segments tokens (``[video]'', ``[caption]'', ``[user1]'', ``[user2]''). We initialize our model with pre-trained GPT2 and introduce three tasks to fine-tune our model: \textbf{Response Language Modeling (RLM)}, \textbf{Video-Audio Sequence Modeling (VASM)}, \textbf{Caption Language Modeling (CLM)}.}
    \label{fig:multimodal-model}
\end{figure*}

These existing methods mainly use independent encoders to separately encode different modalities and then exploit the attention mechanism to fusion the representations of different modalities, which can not fully benefit from the joint representation of multi-modalities. 
To tackle the aforementioned problems of existing methods, in this paper, we design a universal multimodal transformer to encode different modalities and generate responses at the same time. Inspired by Bert \cite{devlin2018bert}, GPT2 \cite{radford2019language} and other pre-training works, we use the self-supervised learning method and adopt the multi-task learning (response language modeling, video-audio sequence modeling, and caption language modeling) approach to learn joint representations and generate informative and fluent responses. To improve the textual representation and generation, we initialize our model with the pre-trained GPT2 \cite{radford2019language} model and then fine-tune it.

Our contributions are as follows:
\begin{itemize}
    \item We are the first to use pre-trained natural language generation models in multimodal dialogue generation.
    \item We integrate multimodal features in one encoder and introduce a multi-task leanring method to learn better joint representations and generate more informative responses.
    \item We achieve a state-of-the-art result on Audio-Visual Scene-Aware Dialog (AVSD) Dataset, outperforming existing methods and other teams in DSTC8-AVSD challenge by a large margin.
\end{itemize}

\section{Related Work}
Most works in the dialogue system focus on open-domain dialogues or task-oriented dialogues. As in human-to-human conversations, there is always background knowledge. Some recent effort develops dialogue systems that can generate responses grounding on the document or structured knowledge graph \cite{li2019incremental,zhou2018dataset,reddy2019coqa,dinan2018wizard,madotto2018mem2seq}. These systems can generate responses that are either more relevant to background knowledge or make more correct interactions. There are also some works incorporating multimodal information in question answering and dialogues. Visual QA \cite{goyal2017making,agrawal2017vqa} is to answer the given question about the content of an image. Visual dialog \cite{das2017visual} is a task to generate natural responses based on the given image and the dialogue context. These works consider text or images as the background knowledge, whereas in Audio-Visual Scene-Aware Dialog the knowledge is video and audio.

It has been shown that pre-trained language models play an important role in improving the performance of language generation tasks, such as the dialogue system and text summarization. \citet{zhang2019pretraining} proposes a natural language generation model based on BERT to make good use of the pre-trained language model in the encoding and decoding process. \citet{wolf2019transfertransfo} introduces transfer learning to generative data-driven dialogue systems using Generative Pretrained Transformer \cite{radford2019language}. In our work, we extend this transfer learning method to multimodal language generation tasks and propose a self-supervised learning method for better video representation.

\section{Methodology}
In this section, we will describe our approaches to multimodal dialogue systems. We will first introduce the Audio Visual Scene-Aware Dialog (AVSD) task. Then we will present our multimodal dialogue generation model and its training methods.

\subsection{Task Formulation} 
Our goal is to integrate multimodal information, which consists of video, audio and dialog context, to generate informative and fluent responses. 
Formally, let $\mathbf{V}$ and $\mathbf{A}$ represent video and audio respectively. Considering the similarity between the summary and the video caption, we concatenate summary and caption as a whole caption $\mathbf{C} = \{c_1, c_2, \ldots, c_I\}$. We use $\mathbf{U} = \{\mathbf{Q}_1, \mathbf{R}_1, \mathbf{Q}_2, \mathbf{R}_2, \ldots \mathbf{Q}_N, \mathbf{R}_N,\}$ to denote the $N$ turns of dialogue, where $\mathbf{Q}_n$ represent the question $n$ and $\mathbf{R}_n = \{r_{n1}, r_{n2}, \ldots, r_{nm} \}$ represent the response $n$ containing $m$ words.
Therefore, the probability to generate the response $\mathbf{R}_n$ for the given question $\mathbf{Q}_n$ considering video $\mathbf{V}$, audio $\mathbf{A}$, dialogue history $\mathbf{U}_{\textless n}$, and caption $\mathbf{C}$ can be computed as:
\begin{align}
    P(\mathbf{R}_n &| \mathbf{V}, \mathbf{A}, \mathbf{C}, \mathbf{U}_{\textless n}, \mathbf{Q}_n; \theta) = \nonumber \\
    &\prod_{j=1}^{m}P(r_{nj} | \mathbf{V}, \mathbf{A}, \mathbf{C}, \mathbf{U}_{\textless n}, \mathbf{Q}_n, r_{n,\textless j}; \theta)
\end{align}

\begin{figure}[t]
    \centering
    \small
    \includegraphics[width=0.35 \textwidth]{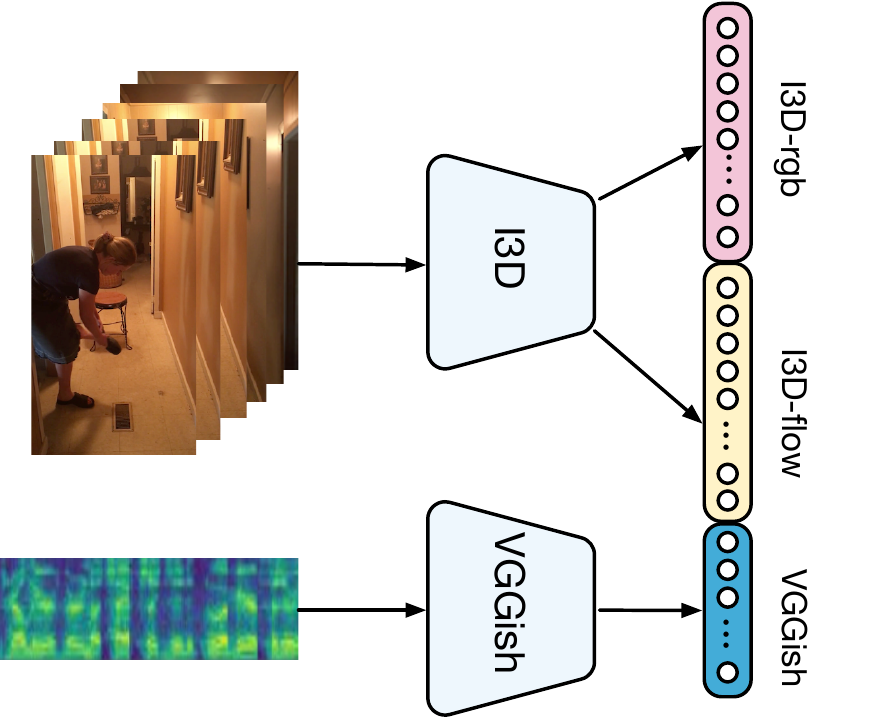}
    \caption{Video and audio feature extractors. For video, we adopt pre-trained I3D-rgb and I3D-flow to extract rgb features and optical flow features. For audio, we use pre-trained VGGish model. }
    \label{fig:video-audio-features}
\end{figure}

\subsection{Model Overview} 
Our model architecture is illustrated in Figure \ref{fig:multimodal-model}, which is a multilayer Transformer encoder based on the GPT2 architecture \cite{radford2019language}. More specifically, we employed a 12-layer decoder-only transformer with the multi-head self-attention.

\noindent \textbf{Text Input}. For text features, we follow GPT2 \cite{radford2019language} and tokenize the input sentence into WordPieces \cite{wu2016google}. 
 
\noindent \textbf{Video and Audio Input}.  For the given video $V_k$, we split the video to $T_k$ segments with a sliding window of $l$ video frames. As shown in Figure \ref{fig:video-audio-features}, for each segment $S_t = \{f_1, f_2, \ldots, f_l\}$, where $f_i$ represents one frame, we use pre-trained I3D-rgb and I3D-flow model to extract $d_v$-dimensional video features $\mathbf{V}_{rgb}$ and $\mathbf{V}_{flow}$. Considering audio is synchronous with video, we select the audio from the same segment and use pretrained VGGish model to extract $d_a$-dimensional audio features as $\mathbf{A}_{vggish}$. We then concatenate video I3D-rgb features, I3D-flow features, and VGGish features:
\begin{equation}
	\mathbf{VA}_t = [\mathbf{V}_{rgb}, \mathbf{V}_{flow}, \mathbf{A}_{vggish}], \mathbf{VA}_t \in \mathbb{R}^{2d_v+d_a}
\end{equation}
Then video-audio features $\mathbf{VA}$ are fed into a fully-connected layer (Video Embedder), as shown in Figure \ref{fig:multimodal-model}, and projected to the same embedding space as text embedding.

\noindent As shown in Figure \ref{fig:multimodal-model}, to make our model have the ability to distinguish among different part of the input (video, caption, speaker1 and speaker2) and make use of the order of the sequence, the final representation for each word token is obtained via summing up its word embedding (WE), positional encoding (PE) and segment embedding (SE). Note that ``[video]'', ``[cap]'', ``[user1]'', and ``[user2]'' are used to represent the segment of video, caption, speaker1, and speaker2 respectively.

\subsection{Multi-task Learning}
We introduce three tasks to fine-tune our model: Response Language Modeling \textit{conditioned on video, audio, caption and dialogue history}, Video-Audio Sequence Modeling \textit{conditioned on caption and dialogue}, and Caption Language Modeling \textit{conditioned on video and audio}. \\

\noindent \textbf{Response Language Modeling (RLM)}. The goal of this task is to generate responses $\mathbf{R}_n = \{r_{n1}, r_{n2}, \ldots, r_{nm}\}$ based on the video-audio features $\mathbf{VA}$, caption $\mathbf{C}$, dialogue history $\mathbf{U}_{\textless n}$, and question $\mathbf{Q}_n$, by minimizing the negative log-likelihood loss function:
\begin{align}
	\mathcal{L}_{RLM}&(\theta) = -E_{(\mathbf{VA}, \mathbf{C}, \mathbf{U}, \mathbf{Q}, \mathbf{R}) \sim D} \nonumber \\
	&\log \prod_{j=0}^{m}P(r_{nj} | \mathbf{VA}, \mathbf{C}, \mathbf{U}_{\textless n}, \mathbf{Q}_n, r_{n,\textless j})
\end{align}
where $\theta$ is the trainable parameters and ($\mathbf{VA}, \mathbf{C}, \mathbf{U}, \mathbf{Q}$) pairs are sampled from the whole training set $D$.

\noindent \textbf{Video-Audio Sequence Modeling (VASM)}. This task is to predict video-audio features given caption and dialogue history. Unlike textual tokens which are represented as discrete labels, video-audio features are high-dimensional and continuous. Instead of clustering video-audio features to discrete labels as \citet{sun2019videobert} do, we adopt the video-audio feature regression method following \cite{chen2019uniter}. This task regresses the Transformer output of video-audio feature $\mathbf{o}_t$ to the next video-audio feature $\mathbf{VA}_{t+1}$. In particular, we apply a fully-connected layer to transform the output to a vector $g_{\theta}(\mathbf{o}_t)$ of the same dimensional as $\mathbf{VA}_{t+1}$. We train this task by minimizing L2 loss:
\begin{align}
	\mathcal{L}_{VASM}(\theta) = &E_{(\mathbf{VA}, \mathbf{C}, \mathbf{U})\sim D} \nonumber \\
	&\frac{1}{T}\sum_{t=1}^{T}\|g_{\theta}(\mathbf{o}_t) - \mathbf{VA}_{t+1}\|^2_2
\end{align}
where $\mathbf{o}_t = f_{\theta}(\mathbf{VA}_{\textless t+1}, \mathbf{C}, \mathbf{U})$ and $f_{\theta}$ represents our model.

\noindent \textbf{Caption Language Modeling (CLM)}. Similar to Response Language Modeling task, we train the model to generate caption $\mathbf{C}=\{c_1, c_2, \ldots, c_I\}$ given the video-audio feature $\mathbf{VA}$:
 \begin{align}
	\mathcal{L}_{CLM}(\theta) &= -E_{(\mathbf{VA}, \mathbf{C}) \sim D} \log \prod_{i=0}^{I}P(c_{i} | \mathbf{VA}, c_{\textless i})
\end{align}

\begin{table*}[tb]
		\centering
		\begin{tabular}{l@{~~~~~~} c@{~~~~~~~}c@{~~~~~~~}c@{~~~~~~~} c@{~~~~~~~}c@{~~~~~~~} c@{~~~~~~~}c }
			\toprule[1pt]
            \bf Models & \bf BLEU-1 & \bf BLEU-2 & \bf BLEU-3 & \bf BLEU-4 & \bf METEOR & \bf ROUGE-L & \bf CIDEr \\ 
            \hline
			\multicolumn{8}{c}{\it Input: text only} \\
			\hline
            Hierarchical Attention  & - ~& - ~& - ~& 0.376  ~&  0.264 ~& 0.554 ~&  1.076 \\
			\bf Our model (RLM) & \bf 0.747 ~& \bf 0.627  ~& \bf 0.527 ~& \bf 0.445  ~& \bf 0.287 ~& \bf 0.594 ~&  \bf 1.261 \\
			\hline
			\multicolumn{8}{c}{\it Input: text + video} \\
			\hline
			Hierarchical Attention  & - ~& -  ~& - ~& 0.394  ~&  0.267 ~& 0.563 ~&  1.094 \\
			MTN  & -  ~& -  ~&  -  ~& 0.392  ~&  0.269 ~& 0.559  ~&  1.066 \\
			\bf Our model (RLM) & 0.759  ~& 0.635  ~&  0.533 ~& 0.448  ~&  0.293 ~& 0.602 ~&  1.282 \\
            ~~~ \bf + VASM & \bf 0.765  ~& \bf 0.643  ~& \bf 0.543 ~& \bf 0.459  ~& \bf 0.294 ~& \bf 0.606  ~&  \bf 1.308 \\
			\hline
			\multicolumn{8}{c}{\it Input: text + video w/o caption / summary} \\
			\hline
			Naive fusion  & -  ~& -  ~&  -  ~& 0.309  ~&  0.215 ~& 0.487  ~& 0.746\\
			DSTC7-AVSD Team 9  & - ~& -  ~&  - ~& 0.315  ~&  0.239 ~& 0.481 ~&  0.773 \\
			\bf Our model (RLM) & \bf 0.694 ~& \bf 0.570  ~& \bf 0.476 ~& \bf 0.402  ~& \bf 0.254 ~& \bf 0.544 ~&  \bf 1.052 \\
			~~~ \bf + VASM & 0.677 ~& 0.556  ~& 0.462 ~& 0.389  ~& 0.250 ~& 0.533 ~&  1.004 \\
			~~~ \bf + recaption & 0.670 ~&  0.537 ~& 0.438 ~& 0.362  ~& \bf 0.254 ~& 0.535  ~& 1.022\\
			
			\bottomrule[1pt]
		\end{tabular}
		\caption{Objective evaluation results on the test set provided by the organizers in DSTC7-AVSD challenge (6 groundtruth available). }
		\label{tab:objective-evaluation-DSTC7}
	\end{table*}
	
\begin{table*}[tb]
		\centering
		\begin{tabular}{l@{~~~~} c@{~~~~}c@{~~~~}c@{~~~~} c@{~~~~}c@{~~~~} c@{~~~~} c@{~~~~}c }
			\toprule[1pt]
            \bf Models & \bf BLEU-1 & \bf BLEU-2 & \bf BLEU-3 & \bf BLEU-4 & \bf METEOR & \bf ROUGE-L & \bf CIDEr & \bf Human rating\\ 
            \hline
			\multicolumn{8}{c}{\it Input: text only} \\
			\hline
			\bf Our model (RLM)  &  0.744 ~& 0.626  ~& 0.525 ~&  0.442  ~& \bf 0.287 ~& 0.595 ~&  1.231 ~& \bf 3.934\\
			\hline
			\multicolumn{8}{c}{\it Input: text + video} \\
			\hline
			\bf Our model (RLM)  &  0.739 ~& 0.624  ~& 0.528 ~& \bf 0.447  ~& 0.284 ~& 0.592 ~&  1.226 ~&  3.895 \\
			
			~~~ \bf + VASM &  \bf 0.746 ~& \bf 0.626  ~& \bf 0.528 ~& 0.445  ~& 0.286 ~& \bf 0.598 ~&  \bf 1.240 ~&  - \\
			\hline
			\multicolumn{8}{c}{\it Input: text + video w/o caption / summary} \\
			\hline
			\bf Our model (RLM)  &  \bf 0.677 ~& \bf 0.556  ~& \bf 0.462 ~& \bf 0.387  ~& \bf 0.249 ~& \bf 0.544 ~&  \bf 1.022 ~&  - \\
			~~~ \bf + VASM &  0.669 ~& 0.550  ~& 0.457 ~& 0.385  ~& 0.246 ~& 0.540 ~&  0.988 ~&  - \\
			~~~ \bf + recaption & 0.661 ~& 0.533  ~& 0.437 ~& 0.364  ~& 0.242 ~& 0.533 ~& 0.991 ~& - \\
			\bottomrule[1pt]
		\end{tabular}
		\caption{Objective and subjective evaluation results on the test set provided by the organizers in DSTC8-AVSD challenge (6 groundtruth available). }
		\label{tab:objective-evaluation-DSTC8}
	\end{table*}
	
	\begin{table*}[tb]
		\centering
		\begin{tabular}{c@{~~~~} c@{~~~~}c@{~~~~}c@{~~~~} c@{~~~~}c@{~~~~} c@{~~~~} c }
			\toprule[1pt]
            \bf History Length & \bf BLEU-1 & \bf BLEU-2 & \bf BLEU-3 & \bf BLEU-4 & \bf METEOR & \bf ROUGE-L & \bf CIDEr\\ 
            \hline
            \bf 0  & 0.729   ~& 0.599  ~& 0.496 ~& 0.413  ~& 0.275 ~& 0.573  ~&  1.182 \\
			\bf 1  & 0.760   ~& 0.638  ~& 0.536 ~& 0.452  ~& \bf 0.296 ~& 0.605  ~&  1.305 \\
			\bf 2  & 0.755   ~& 0.632  ~& 0.532 ~& 0.450  ~& \bf 0.296 ~& 0.601  ~&  1.297 \\
			\bf 3  & \bf 0.765  ~& \bf 0.643  ~& \bf 0.543 ~& \bf 0.459  ~& 0.294 ~& \bf 0.606  ~&  \bf 1.308 \\
			\bf 5  & 0.758  ~& 0.634  ~& 0.533 ~& 0.451  ~& 0.292 ~& 0.601  ~&  1.293 \\
			\bf 9  & 0.759  ~& 0.631  ~& 0.526 ~& 0.441  ~& \bf 0.296 ~& 0.603  ~&  1.294 \\
			\bottomrule[1pt]
		\end{tabular}
		\caption{Objective evaluation results on the test set of DSTC7-AVSD (6 groundtruth available) in which maximum history dialogue turn length ranges from 0 to 3 or 5 or 9. Best result in each metric is highlighted in bold.}
		\label{tab:history}
	\end{table*}
\begin{table*}[tb]
		\centering
		\begin{tabular}{c@{~~~~} c@{~~~~}c@{~~~~}c@{~~~~} c@{~~~~}c@{~~~~} c@{~~~~} c }
			\toprule[1pt]
            \bf Decoding Methods & \bf BLEU-1 & \bf BLEU-2 & \bf BLEU-3 & \bf BLEU-4 & \bf METEOR & \bf ROUGE-L & \bf CIDEr\\ 
            \hline
			\bf Greedy Search  & 0.743   ~& 0.610  ~& 0.503 ~& 0.416  ~& 0.284 ~& 0.587  ~&  1.217 \\
			\bf Nucleus Sampling  & 0.680  ~& 0.525  ~& 0.410 ~& 0.321  ~& 0.252 ~&0.527  ~&  0.955 \\
			\bf Beam Search & \bf 0.765  ~& \bf 0.643  ~& \bf 0.543 ~& \bf 0.459  ~& \bf 0.294 ~& \bf 0.606  ~& \bf 1.308 \\
			\bottomrule[1pt]
		\end{tabular}
		\caption{Objective evaluation results compared between different decoding methods. }
		\label{tab:decoding-methods}
	\end{table*}
\subsection{Different Settings}
\noindent \textbf{Text-only}. We only use text input and Response Language Modeling (RLM) task. \\
\noindent \textbf{Text + Video}. We use text input and video-audio input and we train the model with the aforementioned three tasks.\\
\noindent \textbf{Text + Video w/o caption}. In this setting, there are two methods: 1) Don't use caption in both training and testing, and train the model with Response Language Modeling (RLM) and Video-Audio Sequence Modeling (VASM). 2) Using captions and training the model with the three aforementioned tasks. When testing, first generate video caption based on the given video-audio input (recaption) and then generate responses using video-audio input, generated caption, and dialogue history.\\

\section{Experiments}

\subsection{Datasets}
We use the Audio-Visual Scene-Aware Dialog (AVSD) dataset \cite{alamri2018audio}. In this dataset, each dialog consists of a sequence of questions and answers about the given video between two speakers. There is a video description for each video. We use the state-of-the-art video feature extractor I3D model pre-trained on YouTube videos and the Kinetics dataset \cite{kay2017kinetics}. Specifically, we use the output from the ``Mixed 5c" layer of the I3D network, which is a 2048-dimensional vector. For audio features, we adopted the VGGish model \cite{hershey2017cnn} which outputs a 128-dimensional embedding. There are 7,659 dialogues for training, 1787 dialogues for validation and 1710 dialogues for testing.

\subsection{Baselines}
We compare our model with several related baseline methods:

\noindent \textbf{Naive Fusion}:  The multimodal baseline provided by the organizers, which combines all modalities with a projection matrix \cite{hori2019end}. \\
\textbf{Hierarchical Attention}: The hierarchical attention approach to combine textual and visual modalities, which the team ranked 1$^{st}$ in the DSTC7-AVSD task adopted. \\
\textbf{MTN}: The state of the art system before the DSTC8-AVSD Challenge, which proposes Multimodal Transformer Networks (MTN) to encode videos and incorporate information from different modalities \cite{le-etal-2019-multimodal}. \\

\subsection{Metrics}
\textbf{Objective evaluation}: We report the metrics that are commonly used in the natural language generation tasks, such as BLEU~\cite{papineni2002bleu}, METEOR~\cite{denkowski:lavie:meteor-wmt:2014}, ROUGE-L~\cite{lin2004rouge}, and CIDEr~\cite{vedantam2015cider}. We evaluate our models using the toolkit provided by the competition organizers.

\noindent \textbf{Subjective Evaluation}: Subjective evaluations are essential for dialogue generation. The organizers also evaluate some systems based on crowd-sourced human ratings. The annotators are asked to consider the correctness, naturalness, informativeness, and appropriateness of the generated responses and give a score at 5 levels.

\subsection{Experiment Settings}
In our experiment, we initialize our model with the pre-trained weights from the GPT2-base model \cite{radford2019language,Wolf2019HuggingFacesTS}. In the training process, we use up to 3 turns of history. The hidden size of our model is 768 and the batch size is 32. We use Adam optimizer with a learning rate of 6.25e-5. In the decoding process, we use beam search with a beam size of 5, max length of 20 and a length penalty of 0.3. 

\subsection{Experimental Results}

In this section, we report the experimental results under different settings: text only, text + video, text + video without caption/summary.

\noindent \textbf{Text only}. As shown in Table \ref{tab:objective-evaluation-DSTC7}, compared to Hierarchical Attention which is used in the winner system of DSTC7-AVSD challenge, our model gets better performance on all metrics. In detail, our model improves BLEU-4 by 0.069 and CIDEr by 0.185. Additionally, Table \ref{tab:objective-evaluation-DSTC8} also shows the human evaluation rating in the DSTC8-AVSD track. During human evaluation, the evaluators are asked to rate even the groundtruth references, which are scored 4.000, our model for this task scores 3.934, which is the highest Human rating among all DSTC8 submissions. In the human rating perspective, the results of our model are very close to human dialogue.
	
\noindent \textbf{Text + video}. This task uses full information of the track: caption, summary, dialogue history, and video. As we can see in Table \ref{tab:objective-evaluation-DSTC7}, compared to MTN which is the former state-of-the-art model for this task, our model also achieves a huge improvement. In particular, our model improved BLEU4 by 0.056, and CIDEr by 0.216. Compared to the text-only task, our models achieve better results, which indicates that our method for video understanding is effective. We adopt multi-task learning as we described before. Video-Audio sequence modeling task improves the score of BELU-4 by 0.011 and CIDEr by 0.026. In DSTC8 results shown in Table \ref{tab:objective-evaluation-DSTC8}, this method get a little lower BELU4 value, but still improved CIDEr by 0.014, which shows the method is effective. 

\noindent \textbf{Text + video w/o caption/summary}. This setting is most similar to the actual scene-aware dialogue: we only have video-audio information and dialogue history. Therefore, this task is more challenging. As shown in Table  \ref{tab:objective-evaluation-DSTC7},  we can see the lower performance than the text + video
task as expected, but it is gratifying that our model still got a relatively high performance. We outperform the DSTC7-AVSD Team 9 who got the highest performance in this task by a large margin. In this task, we also try to use the multi-task learning method but gets lower performance on almost all metrics. We will discuss this phenomenon in the next section. 

\subsection{Analysis and Discussion}

\noindent \textbf{Training Method Analysis}. As we introduced in Experiment Results, after we adopt the feature regression, our model gets better performance in text + video task, but get lower performance in text + video w/o caption/summary task. We think the reason is the shortage of information: only uses the history dialogue, it is difficult to rebuild the masked video feature, and compared to rebuilding text from adjacent context, we think our model doesn't have a strong ability in extracting information from videos, therefore, the method doesn't work in this task. 

For the poor performance of CLM, we think the reason may be similar: the poor ability in extracting video information of our model limited the performance for inferring caption from the video. So we think future work can focus on video comprehension as well as integrating video and text information.

\noindent \textbf{History Length}. 
We experiment with our model in text + video settings with video regression loss to explore the influence of dialogue history length. As shown in Table \ref{tab:history}, our model gets the best performance when the maximum dialogue history length is 3.

\noindent \textbf{Decoding Methods}. To find an effective decoding method for multimodal dialogue generation, we try various decoding methods, including greedy search, beam search, and nucleus sampling \cite{holtzman2019curious} which samples text from the dynamic nucleus of the probability distribution. As shown in Figure \ref{tab:decoding-methods}, decoding with beam search gets the best results on all objective results among these three decoding methods. We consider that in Audio-Visual Scene-Aware Dialog, grounding on video and caption, responses are relatively more definite than that in open-domain dialogues. Therefore, it's better to use the beam search when decoding in this task.


\section{Conclusion and Future Work}
In this paper, we propose a multimodal dialogue generation model based on a pre-trained language model and introduce multi-task learning to learn more accurate joint representation among multi modalities and generate more informative responses. 
In the future, we plan to use more video features like ResNet features and explore more training tasks to improve the joint understanding of video and text. In addition, we hope to extend these methods to other tasks, such as video captioning, image captioning, and visual dialog.

\section{Acknowledgments}
 We sincerely thank the anonymous reviewers for their thorough reviewing and valuable suggestions. This work is supported by National Natural Science Foundation of China (NO.61876174) and National Key R\&D Program of China (NO.2017YFE9132900).

\bibliography{aaai-2020}
\bibliographystyle{aaai}
\end{document}